\documentclass[10pt,twocolumn,letterpaper]{article}

\usepackage[pagenumbers]{cvpr} % To force page numbers, e.g. for an arXiv version

\usepackage{graphicx}
\usepackage{amsmath}
\usepackage{amssymb}
\usepackage{booktabs}
\usepackage{subcaption}
\usepackage{enumitem}

\usepackage[pagebackref,breaklinks,colorlinks]{hyperref}

\usepackage[capitalize]{cleveref}
\crefname{section}{Sec.}{Secs.}
\Crefname{section}{Section}{Sections}
\Crefname{table}{Table}{Tables}
\crefname{table}{Tab.}{Tabs.}

\def\cvprPaperID{11789} % *** Enter the CVPR Paper ID here
\def\confName{CVPR}
\def\confYear{2022}

\begin{document}

\title{A Framework for Learning Ante-hoc Explainable Models via Concepts}

\author{Anirban Sarkar\\
IIT Hyderabad\\
Telangana, India\\
{\tt\small cs16resch11006@iith.ac.in}
\and
Deepak Vijaykeerthy\\
IBM Research\\
Bangalore, India\\
{\tt\small deepakvij@in.ibm.com}
\and
Anindya Sarkar\\
IIT Hyderabad\\
Telangana, India\\
{\tt\small anindyasarkar.ece@gmail.com}
\and
Vineeth N Balasubramanian\\
IIT Hyderabad\\
Telangana, India\\
{\tt\small vineethnb@iith.ac.in}
}

\maketitle

%%%%%%%%% ABSTRACT
\begin{abstract}
\vspace{-10pt}
   Self-explaining deep models are designed to learn the latent concept-based explanations implicitly during training, which eliminates the requirement of any post-hoc explanation generation technique. In this work, we propose one such model that appends an explanation generation module on top of any basic network and jointly trains the whole module that shows high predictive performance and generates meaningful explanations in terms of concepts. Our training strategy is suitable for unsupervised concept learning with much lesser parameter space requirements compared to baseline methods. Our proposed model also has provision for leveraging self-supervision on concepts to extract better explanations. However, with full concept supervision, we achieve the best predictive performance compared to recently proposed concept-based explainable models. We report both qualitative and quantitative results with our method, which shows better performance than recently proposed concept-based explainability methods. We reported exhaustive results with two datasets without ground truth concepts, i.e., CIFAR10, ImageNet, and two datasets with ground truth concepts, i.e., AwA2, CUB-200, to show the effectiveness of our method for both cases. To the best of our knowledge, we are the first ante-hoc explanation generation method to show results with a large-scale dataset such as ImageNet.
   
%   Self-explainable deep models are devised to represent the hidden concepts in the dataset without requiring any posthoc explanation generation technique. Here we worked with one of such models motivated by adding an explanation generation module on top of a basic network and jointly training the whole module that not only shows high predictive performance but also generates meaningful explanations in terms of concepts. We proposed a training strategy by applying self-supervised training on the concepts to extract better explanations. We showed that the model can be further improved by leveraging full concept supervision which attains the best predictive performance compared to recently proposed concept based explainable models. Our proposed method is very simple to implement and generates better explanation in terms of various metrics. We reported exhaustive results with two datasets without ground truth concepts i.e. CIFAR10, ImageNet and two other datasets with ground truth concepts i.e. AWA2, CUB to show effectiveness of our method for both cases. To the best of our knowledge, we are the first one to show ante-hoc explanation generation for a large scale dataset such as ImageNet.

\end{abstract}

%%%%%%%%% BODY TEXT
\vspace{-10pt}
\section{Introduction}
\vspace{-5pt}
\label{sec:intro}
% * Overview of explainable models (in vision) and convey how most of the work is in post-hoc explainability --> end this para with motivating ante hoc explainable
Recent years have seen an exponentially increasing interest in explainability of decisions of Deep Neural Network (DNN) models across domains including biometrics, healthcare, autonomous navigation and many more. Existing efforts in computer vision including occlusion-based, gradient-based and Shapley value-based efforts largely perform post hoc analysis~\cite{cam, grad-cam, attr2016perturb_3_lime}, of an already trained model to identify what a DNN model looked at in an input image while making a prediction. While this is useful, the separation of explanation from prediction is not ideal.  When an explanation goes wrong, it is not trivial to understand if the explanation method is incorrect, or if the model itself relied on spurious correlations to make a prediction. This has paved the need for ante hoc methods that jointly learn to explain and predict, and thus learn inherently interpretable models.

\begin{figure}
  \centering
  \includegraphics[scale=0.5]{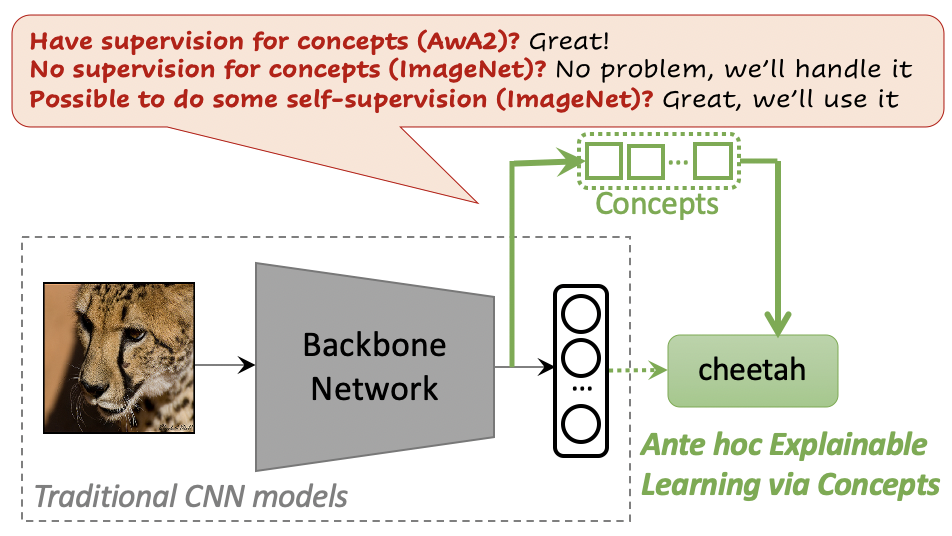}
  \caption{Illustration of the proposed framework. Our framework offers a way to train models that can not only predict but also explain their predictions. It can be easily integrated with existing backbone networks. Compared to existing techniques, it provides the flexibility to incorporate different forms of supervision (including weaker forms of supervision like \textit{self-supervision}) whenever available or feasible.}
  \label{fig:introduction}
\end{figure}

%     * An overarching summary of what ante hoc explainability methods exist, and where they fall short --> ideally motivate our method
Efforts on envisioning interpretable learning models by Rudin \cite{rudin2019stopexplaining} and Lipton \cite{lipton2018mythos} have stressed on the importance of implicitly interpretable methods over post hoc explanations in elaborate terms. In a more recent exposition, Rudin et al \cite{rudin2021interpretableml} identified ten challenges of interpretable machine learning, which also highlighted the need for placing constraints into models to learn with better interpretability during training itself. 
The last few years have seen a few efforts in ante hoc methods that explain through concepts, which are learned during the training of the DNN itself such as Self-Explaining Neural Networks \cite{alvarez2018towards},
Concept Bottleneck Models \cite{koh2020concept}, Concept-based Model Extraction \cite{kazhdan2020now}, and 
Concept Whitening \cite{chen2020concept}. Learning concepts during training provides a natural pathway for ante hoc explanations that are global (concepts that are most activated on a dataset or a class) or local (concepts that are most activated for prediction on given input image). Existing methods however either require concept-level supervision to train the model \cite{koh2020concept}, or require a significant number of additional parameters in the network \cite{alvarez2018towards}, which prohibits their use in deeper models more commonly used in practice. %In particular, they have not been very efficient or cannot operate with different forms of supervision or weaker supervision to learn ante hoc concept-based explanations of DNN model predictions.

%     * Usefulness of ante hoc methods (cite Cynthia Rudin's work) -- (i) you can reason in terms of concepts; (ii) such an approach can help handle corruptions at test time; and (iii) can leverage different levels of supervision, based on what is available
In this work, we propose a new method towards learning ante-hoc explanations via concepts, that: (i) can be added easily to existing backbone classification architectures with minimal  additional parameters; (ii) can provide explanations for model decisions in terms of concepts for an individual input image or for groups of images; and (iii) can work with different levels of supervision, including \textit{no concept-level supervision} at all. This is achieved by an architectural modification added to a backbone network along with additional loss terms that allow such ante hoc learning. Importantly, we show that our framework allows learning of concepts with no supervision, self-supervision as well as full supervision at a concept-level. An overview of our proposed model is shown in fig.\ref{fig:introduction}.

%     * List of contributions (Very simple and efficient)
Our key contributions in this work can be summarized as follows:
\vspace{-7pt}
\begin{itemize}[leftmargin=*]
\setlength\itemsep{-0.25em}
    \item We propose a simple and effective method that jointly learns to predict and explain (through concepts) in an ante hoc manner (i.e. learning to explain during training itself, as opposed to post hoc explainability methods popularly used today).
    \item Our method can learn to explain through concepts with different levels of supervision: (i) with no concept-level supervision;  (ii) through weak supervision (self-supervised learning of concepts); as well as (iii) with concept-level supervision.
    \item We perform a comprehensive suite of experiments to study accuracy and explainability of our method on multiple benchmark datasets quantatively and qualitatively, and show ablation studies on different choices made in the method. In this context, we introduce a metric based on concept intervention for ante hoc explainable models such as ours.
    \item Our method outperforms existing methods on accuracy and explainability metrics, and achieves these results with negligible computational overhead over baseline models with no explanation component.
\end{itemize}

\section{Related Work}
\vspace{-5pt}
\label{sec:related}
The main objective of concept learning is to obtain a lower-dimensional representation that faithfully explains the downstream tasks, such as - object classification. 

% In general, such a low dimensional representation can be characterized as $C \in \mathbb{R}^{K \times d}$, i.e. every concept $c \in \mathbb{R}^d$ belongs to one of the total $k \in K$ concepts.

% In our work, we learn one-dimensional concepts, i.e., our setup uses $k$ concepts, with every concept is represented by a scalar value. 
% A recent class of unsupervised and supervised concept learning methods such as \cite{alvarez2018towards,koh2020concept,yeh2019completeness,kazhdan2020now} follow a similar notion of concepts.
% Like most other concept learning methods, in this work, we assume $d = 1$, or in other words, we represent a concept by 1 dimension. In this setup, an element of the j-th column is expected to have a high value if the j-th concept is considered to be activated for the input that generated this representation. In this work, we make use of a concept encoder that maps a transformed input to a intermediate low level concept representation, which is in line with the previous works, such as \cite{koh2020concept,yeh2019completeness,kazhdan2020now}.

\noindent \textbf{Unsupervised Concept Learning}: Most of the existing methods generate meaningful explanations in an unsupervised manner, i.e., when ground truth concepts are not available for the dataset. Such methods either work as a post-hoc approach on a trained model \cite{kim2018interpretability} or learn an inherently interpretable model \cite{alvarez2018towards,yeh2019completeness}. TCAV \cite{kim2018interpretability} leverages the directional derivatives with intermediate model features to quantify the importance of a user-defined concept towards final model predictions. Though this method doesn't require full concept annotations, the explanations are generated based on the prior knowledge of concepts over the data points. Zhou et al. \cite{zhou2018interpretable} proposed a method to decompose the model prediction in terms of projections onto concept vectors using the model generated saliency by CAM \cite{zhou2016learning}. Another recent method \cite{yeh2019completeness} leverages Shapley values to quantify the sufficiency of a set of concepts in explaining the model predictions through completeness measure of the concepts. Being a post-hoc explainability method, it works on trained deep networks. Unlike our method, it doesn't allow a user to intervene on the concepts to explore the interactions between concepts and the class predictions.

The first fully-unsupervised ante-hoc concept learning method, SENN \cite{alvarez2018towards}, employs a concept encoder $h(x)$ with corresponding relevances $\theta(x)$ for an image $x$ and outputs the final logit as $\theta(x)^Th(x)$. SENN is trained, following a joint training approach, with cross-entropy loss for the logits and a stability loss to enforce closeness of the similar concept relevances i.e. $\theta(x)$. Similar to SENN, our method also uses a concept encoder to extract concepts. But, we replace the heavy relevance network with a couple of simple, fully connected networks that generate explanations and perform classification.

\noindent \textbf{Supervised Concept Learning}: Methods such as concept bottleneck models (CBM) \cite{koh2020concept} divides the complete model into two parts. The first part is a function $g:X\rightarrow C$, that generates an intermediate concept representation $c$ from an image $x$, which is followed by the label predictor part $f:C\rightarrow Y$ to output a class label from $c$. The model predicts a class label for an image $x$ by computing $f(g(x))$. This model is trained with both concept and class label supervision, either training individual parts sequentially or both parts jointly. Kazhdan et al. proposed CME \cite{kazhdan2020now}, a post-hoc data-efficient version of CBM, that captures intermediate representations from a pre-trained model to improve the sensitivity to the dependence between the concepts and the final prediction. Concept whitening (CW) \cite{chen2020concept} proposed a method to plug an intermediate layer in place of the batch-normalization layer of any pre-trained CNN model that helps in concept extraction by constraining the latent layer output to represent a target concept. As opposed to CBM, we decouple the process of generating explanations and predictions. This helps us to learn concept-based explanations without losing much in predictive performance and enables the user to use the model with different levels of supervision. 

\noindent \textbf{Self-Supervised Concept Learning}:
Different self-supervised methods have been proposed to help learn better representations and boost classification accuracy. Tasks such as predicting the relative position of image patches \cite{doersch2015unsupervised}, predicting rotation angle \cite{gidaris2018unsupervised}, recovering color channels \cite{zhang2016colorful}, solving jigsaw puzzle games \cite{noroozi2016unsupervised}, and discriminating images created from distortion \cite{dosovitskiy2015discriminative} have been extensively used in recent years. Another class of methods reconstruct images from corrupted versions or just part of it such as denoising autoencoders \cite{vincent2008extracting}, image inpainting \cite{pathak2016context}, and split-brain autoencoder \cite{zhang2017split}. Contrastive learning is another paradigm where representations are learned in such a way that similar data points are brought closer, and dissimilar data points are pushed further away \cite{wang2015unsupervised} in the representation space. Predicting natural ordering or topology of data has also leveraged as pretext tasks in video-based \cite{wei2018learning,misra2016shuffle,fernando2017self}, graph-based \cite{hu2019strategies,yang2020self}, and text-based \cite{radford2018improving,devlin2018bert} self-supervised learning.  While self-supervision has been used to learn better model representations, their utility for learning concept-based explanations hasn't been explored in the past. In our work, we explore how self-supervision can be used for learning better concept-based explanations.

% \section{Background}
% \label{sec:background}
% \input{background}

\vspace{-5pt}
\section{Method}
\vspace{-5pt}
\label{sec:method}
\begin{figure*}
  \centering
  \includegraphics[scale=0.44]{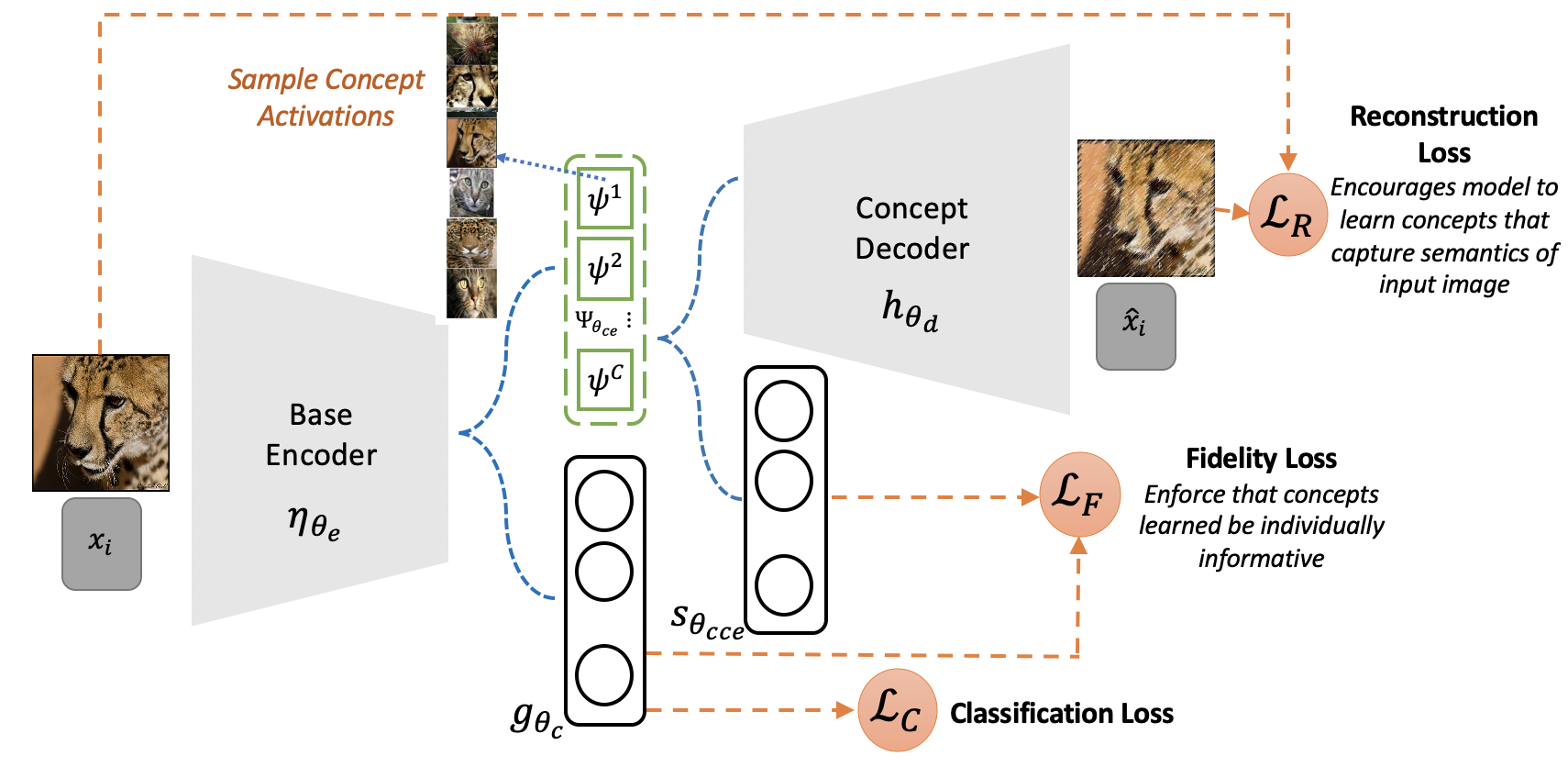}
  \caption{An overview of our proposed framework (concept activations denote images that maximally active each concept)}
\end{figure*}

Let $\mathcal{X}$ denote the input space, and $\mathcal{Y}$ the output space, we assume that the training instances (or examples) $\mathcal{D} = \{x_i, y_i\}_{i = 1}^{N}$ are sampled i.i.d from the source distribution $P$ defined over $\mathcal{X} \times \mathcal{Y}$. We also assume that $\mathcal{X} = \mathbb{R}^d$, and $\mathcal{Y} = \{y \in \{0, 1\}^M, \sum_{j = 1}^M y^k = 1\}$, where $M$ is the number of classes, and $y$ is an one-hot encoded vector. 

We propose a generic framework to incorporate \textit{ante-hoc explanation (or self-explanation) modules} into existing deep learning pipelines. In this paper we demonstrate it for a classification task. In practice, for classification tasks we learn a Deep Neural Network $f_{\theta} = \{\eta_{\theta_{e}}(.), g_{\theta_c}(.)\}$ which consists of a base encoder (or a feature extractor) $\eta_{\theta_{e}(.)}$, that extracts the representation the representation vectors which are fed into a classifier function $g_{\theta_c}(.)$ (a classifier function takes the latent representation $\mathbf{z} = \eta_{\theta_{e}}(x_i)$, and then predicts the label). Typically the base encoder \& the classifying function are trained together by optimizing for $\theta = \{\theta_e, \theta_c\}$ such that the output of the network $\Tilde{y}_i = f_{\theta}(x_i)$ minimizes a loss $\mathcal{L}_C(\Tilde{y}_i, y_i)$ over the set of training instances $\mathcal{D}$.

To incorporate implicit learning of interpretable concepts, in addition to the existing components of classical classification pipelines described previously, we introduce a concept encoder $\Psi_{\theta_{ce}}(.)$ which takes the representation $\eta_{\theta_e}(.)$ and learns a set of \textit{intepretable concepts} $\{\psi^1, \ldots, \psi^C\}$ (where $C$ is the number of concepts), to explain the predictions provided by $f_{\theta}$. In general, concepts are low dimensional representation that can be characterized as $C \in \mathbb{R}^{K \times d}$, i.e. every concept $c \in \mathbb{R}^d$ belongs to one of the total $k \in K$ concepts. In our work, we learn one-dimensional concepts, i.e., our setup uses $k$ concepts, with every concept is represented by a scalar value.

To encourage the model to learn concepts $\{\psi^1, \ldots, \psi^C\}$ that capture the semantics of the input image $x_i$ we pass the concepts to a decoder $h_{\theta_d}(.)$ which reconstructs the image $\hat{x}_i$. We then add a loss $\mathcal{L}_R(x_i, \hat{x}_i)$ which measures the reconstruction error to the overall loss $\mathcal{L}$. $\mathcal{L}_R$ penalises the model $f_{\theta}$, if the concepts aren't suffice to generate an accurate reconstruction $\hat{x}_i$ of the input image $x_i$.  In our paper, we use an $L_2$ loss.

Since the concepts $\{\psi^1, \ldots, \psi^C\}$ explain the prediction of a DNN $f_{\theta}$. Ideally, they should be informative enough by themself to predict the input instance $x_i$ correctly. To enforce that the learnt concepts not only explain the prediction but are also informative, we penalize the model $f_{\theta}$, if the predictions $s_{\theta_{cce}}(\Psi_{\theta_{ce}}(.))$ (where $s_{\theta_{cce}}$ is a classification function which predicts the class labels taking the concepts as input) based on the concepts $\{\psi^1, \ldots, \psi^C\}$ and prediction by the DNN $f_{\theta}$ differs. We enforce that the concepts learned should be individually informative by adding a fidelity loss $\mathcal{L}_F$ to the overall loss $\mathcal{L}$.

Taking the proposed modifications into consideration, the overall loss $\mathcal{L}_{O}$ of the model can be written as follows:
\begin{equation}
\begin{split}
        \mathcal{L}_O &= \mathcal{L}_C(y_i, \Tilde{y}_i) + \alpha\mathcal{L}_R(x_i, \hat{x}_i) \\
        & + \beta\mathcal{L}_{F}(f_{\theta}(x_i), s_{\theta_{cce}}(\Psi_{\theta_{ce}}(x_i)))
\end{split}
\end{equation}

In practice, most data sets seldom include annotations of concepts (or attributes) that could be used to learn a self-explaining model. However, few exceptions contain concepts (or attributes) that the models can leverage while learning to explain their predictions. The majority of existing frameworks either work when only annotation of concepts is available, or data sets don't contain any additional annotations, but not both. Often it is neither trivial nor efficient to incorporate alternate forms for supervision in these existing frameworks. In comparison, our framework offers the flexibility to incorporate different forms of supervision whenever available easily. To illustrate this, we demonstrate how to incorporate i) complete supervision (supervised learning of interpretable concepts), ii) zero-supervision (unsupervised learning of interpretable concepts), and iii) a weaker form of supervision that is cheaply available like self-supervision.

By default, our framework works with data sets where the annotation of concepts isn't available. In cases when they are available, we can easily incorporate them into the learning process by adding a loss $\mathcal{L}_E(\Psi_{\theta_{ce}}(x_i), a_{x_i})$ (where $a_{x_i}$ is the concept (or attribute) annotation of $x_i$). $\mathcal{L}_E(\Psi_{\theta_{ce}}(x_i), a_{x_i})$ would penalize the model if the concepts learned aren't similar to the annotation in the data set for the corresponding instance. We then train the model by optimising $\theta$ such that the output of the network $\Tilde{y}_i = f_{\theta}(x_i)$ minimizes a loss $\mathcal{L}_O + \mu\mathcal{L}_E$ over the training set.

Even when direct supervision isn't available for concepts, it is possible to learn a robust set of high fidelity interpretable concepts $\{\psi^1, \ldots, \psi^C\}$ by leveraging the underlying structure of the data by incorporating supervisory signals obtained directly from the data itself. This technique is popularly known as self-supervision. In our framework, we incorporate \textit{self-supervision} as an auxiliary task with a loss $\mathcal{L}_{SS}$, and the auxiliary task shares the parameters with our model until the concept encoder $\Psi_{\theta_{ce}}(.)$. In this paper, we choose rotation prediction as an auxiliary task. The task involves rotating the image by one of 0, 90, 180, or 270 degrees and predicting the rotation angle $r_i$ as a four-way classification problem through an auxiliary head. We can also easily incorporate other self-supervision tasks into our framework. 

As opposed to existing techniques where the branch for the auxiliary task uses the output of feature extractor (or base encoder) for the self-supervision tasks, in our case, we use the output of the concept encoder $\Psi_{\theta_{ce}}(.)$. In turn, this helps us to ensure that the set of interpretable concepts from the concept encoder $\Psi_{\theta_{ce}}(.)$ always respects the underlying structure of the data and has high fidelity. To estimate $\mathcal{L}_{SS}$ we pass the output of the concept encoder $\Psi_{\theta_{ce}}(.)$ through a classifier function $\zeta_{\theta_{ss}(.)}$ that predicts the angle of rotation, we then compute cross-entropy between $\zeta_{\theta_{ss}(.)}$ and $r_i$. Like in other cases, we jointly train the model and the auxiliary head by optimizing $\theta$ such that the output of the network $\Tilde{y}_i = f_{\theta}(x_i)$ minimizes a loss $\mathcal{L}_O + \gamma\mathcal{L}_{SS}$ over the training set. In cases where ground truth annotations of concepts aren't available, and an auxiliary self-supervision task isn't used $\mu$, and $\gamma$ are respectively set to 0.

\[
\mathcal{L}^{'}_O = \mu\mathcal{L}_E(\Psi_{\theta_{ce}}(x_i), a_{x_i}) + \gamma\mathcal{L}_{SS}(r_i, \zeta_{\theta_{ss}(.)})
\]

Even though our framework incorporates additional components to existing deep learning backbones (or pipelines), we can discard most of them after the training. We only retain the sub-network (or module) to generate explanations in addition to the ones on standard deep learning pipelines (i.e., feature extractor and the classifier function) during the prediction time. Hence, compared to existing self-explaining models, the additional cost incurred by our framework is relatively insignificant.

\vspace{-5pt}
\section{Experiments}
\vspace{-5pt}
\label{sec:expts}
We show that our framework achieves competitive predictive accuracy compared to standard classification pipelines, as well as meaningful explanations. We report results with our method on CIFAR10, ImageNet, AwA2 and CUB-200 with different levels of concept supervision according to availability of ground truth concepts i.e. unsupervised manner on CIFAR10 \cite{Kriz2009cifar}, ImageNet \cite{deng2009imagenet} and with concept supervision on AwA2 \cite{lampert2013attribute}, CUB-200 \cite{WelinderEtal2010}. We also report results with AwA2, CUB when our model is trained without concept supervision to show the effectiveness of our method in both cases, i.e., with and without concept supervision. We consider SENN \cite{alvarez2018towards} and CBM \cite{koh2020concept} as our baselines considering the basic methods for unsupervised and supervised concept learning.

\noindent \textbf{Dataset Details}: The CIFAR-10 dataset \cite{Kriz2009cifar} consists of 32x32 colour images in 10 classes, each with 5000 train images and 1000 test images  per class. The ImageNet dataset \cite{deng2009imagenet} is comprised of more than 1 million images and 1,000 object classes of natural images. AwA2 dataset \cite{lampert2013attribute} consists of 37322 images of total 50 animals classes with 85 numeric attribute. The other attribute dataset we considered is CUB-200 \cite{WelinderEtal2010}, an image dataset with photos of 200 bird categories with a total of 6033 images and 312 attribute annotations for each image.

\noindent \textbf{Architecture Details}: We use ResNet18 \cite{he2016identity} as our backbone network for all the datasets, as there is no standard architecture followed in the literature related to concept learning. The backbone network resembles to $f_{\theta} = \{\eta_{\theta_{e}}(.), g_{\theta_c}(.)\}$ as given in Sec.\ref{sec:method}. The output of the feature encoder $\eta_{\theta_{e}}(.)$ is also passed to the concept encoder $\Psi_{\theta_{ce}}(.)$ which is a single fully connected layer, that outputs a set of interpretable concepts $\{\psi^1, \ldots, \psi^C\}$. \textit{Here $C$ is the number of concepts, and we have considered 10 and 100 concepts for CIFAR10 and ImageNet, respectively. The number of concepts (or attributes) for AwA2 and CUB-200 is 85 and 312, respectively.} We have kept the number of concepts the same for a fair comparison while training our model with these datasets for both unsupervised concept learning and learning with concept supervision. The classification function $s_{\theta_{cce}}$ that predicts the class labels, taking the concepts as input, is also a single fully connected layer. The number of parameters of the concept encoder and the classification network, taking the concepts as inputs, vary for different datasets based on the number of concepts and classes. We implement the decoder $h_{\theta_d}(.)$ as a set of deconvolution layers.

\noindent \textbf{Storage and Time Complexity}: The architecture proposed by SENN requires a vast number of parameters with both the concept and relevance encoder contributing to it. Our model alleviates this issue by removing the relevance network altogether and adding the concept classification network that serves a similar purpose. However, the decoder network is required to make the concepts capture sufficient information to reconstruct the image. Hence, our overall network requires $\sim$60\% of the space and training time compared to SENN. Compared to CBM, our method requires $\sim$1.5 times the space and training time. This is due to a decoder that enables our framework to support cases when concept supervision isn't available, which CBM doesn't. Our approach takes almost similar inference time as CBM as we don't use the decoder network during inference and concept extraction. Please note that these storage and time measurements during training are with ResNet18 backbone architectures, and the gap with CBM will further reduce with more complex backbone networks.

\begin{table}
\footnotesize
    \centering
\begin{tabular}{|p{1.45cm}| p{1.2cm} p{1.2cm}|p{1.1cm}|p{0.9cm}|}
		\hline \hline
			& \multicolumn{2}{|c|}{Baselines}&\multicolumn{2}{|c|}{OURS}\\
			Dataset & SENN  & CBM & w/o sup & w sup \\
			\hline \hline
            CIFAR10 & 84.50 & NA & \textbf{91.68} & NA \\
            \hline
            ImageNet  & 58.55 & NA & \textbf{65.09} & NA \\
            \hline
            AwA2     & 76.41 & 81.61 & 81.04 & \textbf{85.70} \\
            \hline
            CUB-200 & 58.81 & 64.17 & 63.05 & \textbf{65.28} \\
            \hline \hline
		\end{tabular}
		\caption{\textit{Accuracy} (in \%) of different methods on CIFAR10, ImageNet, AwA2 and CUB-200 datasets using ResNet18 architecture as concept (or base) encoder}
% 		\vspace{-6pt}
		\label{tab:predictive performance}
\end{table}

\noindent \textbf{Predictive Performance}: Table \ref{tab:predictive performance} reports the predictive performance of our method as well as the baseline methods with CIFAR10, ImageNet, AwA2, and CUB datasets. As CBM requires concept supervision, we can't use this method for CIFAR10 and ImageNet.  An unsupervised version of our method outperforms SENN significantly for all the datasets. CBM, being a method with concept supervision, performs slightly better than our unsupervised version. Our approach, with concept supervision, beats CBM by a large margin. Please note that the predictive performance by our method, reported in table \ref{tab:predictive performance}, is solely based on the backbone network $f_{\theta}(.)$. We decoupled the main prediction task and concept extraction so that our model doesn't sacrifice much of the predictive performance and still can produce meaningful explanations.

\subsection{Quantitative Evaluation}

\begin{figure*}
    \centering
    \includegraphics[scale=0.4]{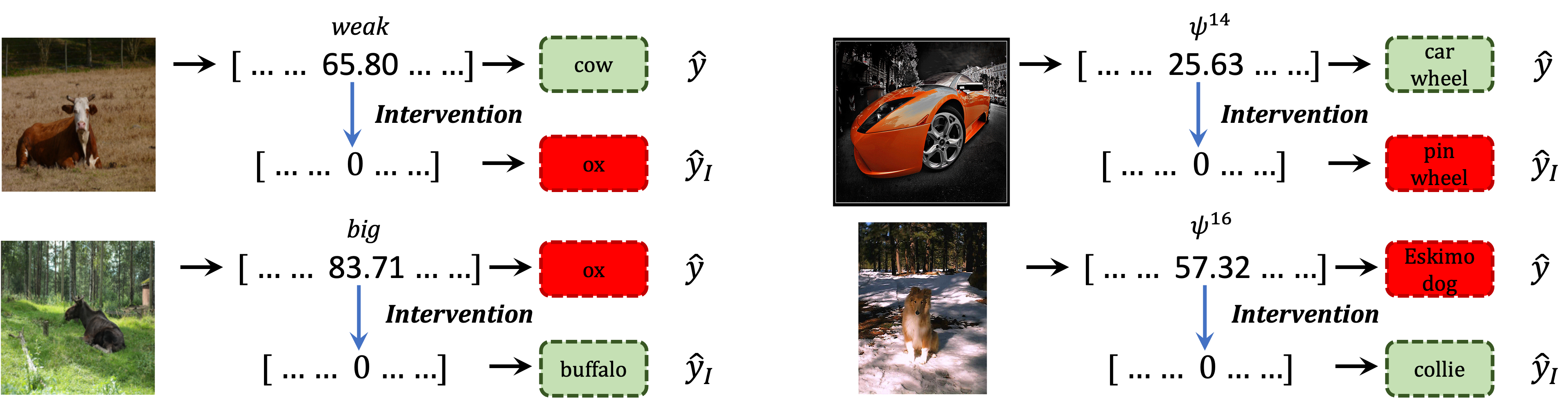}
    % \vspace{-3pt}
    \caption{Successful examples of test-time intervention, where intervening on a single concept changes the model prediction (from $\hat{y}$ to $\hat{y}_{I}$) to the correct label on bottom examples and to the incorrect label on top examples, for AwA2 (left) and ImageNet (right).}
    % \vspace{-8pt}
    \label{fig:interventions}
\end{figure*}

We evaluate and compare the concept-based explanations generated by our method with other state-of-the-art frameworks like SENN and CBM. We consider metrics of interpretability that assess the effectiveness of additional losses we use in our framework. Apart from the existing metrics such as faithfulness, fidelity, and explanation error, we also perform interventions on the generated concepts to illustrate their meaningfulness. Fig.~\ref{fig:interventions} shows examples of interventions that lead to the model changing its prediction when we intervene on the top concept. Besides the predictive performance, our method consistently outperforms the baseline methods in all the other explainability metrics as explained below.

\noindent \textbf{Faithfulness Metric}: In practice, we want the concepts learned to be meaningful and faithfully explain the model's predictions. To evaluate how faithful the explanations generated by different frameworks are, we measure the predictive capacity of the generated concepts, i.e., from the output of $s_{\theta_{cce}}$, in our case. This metric represents the capability of the overall concept vector to predict the ground truth task label. It is similar to other measures such as explicitness \cite{ridgeway2018learning} and informativeness \cite{eastwood2018framework} used to measure feature disentanglement.  
% Whether the overall concept vector can predict the ground truth task label with low prediction error. It is measured by average prediction accuracy through the concept vector.

\begin{table}[h!]
\footnotesize
    \centering
\begin{tabular}{|p{1.45cm}| p{1.2cm} p{1.2cm}|p{1.1cm}|p{0.9cm}|}
		\hline \hline
			& \multicolumn{2}{|c|}{Baselines}&\multicolumn{2}{|c|}{OURS}\\
			Dataset & SENN \cite{alvarez2018towards} & CBM \cite{koh2020concept} & w/o sup & w sup \\
			\hline \hline
            CIFAR10 & 84.50 & NA & \textbf{90.86} & NA \\
            \hline
            ImageNet  & 58.55 & NA & \textbf{59.73} & NA \\
            \hline
            AwA2     & 76.41 & 81.61 & 79.29 & \textbf{83.30} \\
            \hline
            CUB-200 & 58.81 & 64.17 & 61.49 & \textbf{62.59} \\
            \hline \hline
		\end{tabular}
		\caption{Comparison of \textit{faithfulness} (in \%, predictive performance solely based on concepts) of concepts generated by different methods on CIFAR10, ImageNet, AwA2 and CUB-200 data sets.}
% 		\vspace{-6pt}
		\label{tab:effectiveness}
\end{table}

\noindent \textbf{Fidelity Metric}: Fidelity measures the fraction of the data points where the model prediction matches the prediction from the interpretation. It is widely used to measure how well the generated explanations approximate the model\cite{robustexplain}. This metric does not apply to methods where the interpreter is directly used to provide model prediction, such as SENN and CBM. Table \ref{tab:fidelity} reports all the comparative results for all the datasets. We use fidelity loss $\mathcal{L}_{F}$ during training which justifies the high fidelity score for our models. 

% The fraction of samples where prediction of a model and its interpreter agree, i.e predict the same class, is referred to as fidelity. It is a commonly used metric to measure how well an interpreter approximates a model.

\begin{table}[h!]
\footnotesize
    \centering
\begin{tabular}{|p{1.45cm}|p{1.1cm}|p{0.9cm}|}
		\hline \hline
			& \multicolumn{2}{|c|}{OURS}\\
			Dataset & w/o sup & w sup \\
			\hline \hline
            CIFAR10 &  \textbf{99.11} & NA \\
            \hline
            ImageNet  & \textbf{90.22} & NA \\
            \hline
            AwA2     &  \textbf{97.84} & 97.19 \\
            \hline
            CUB-200     &   \textbf{97.52} & 95.87 \\
            \hline
            \hline
		\end{tabular}
		\caption{Comparison of \textit{fidelity} (the \% of match between model prediction and the prediction through the interpretation.) of concepts generated by different methods on CIFAR10, ImageNet, AwA2 and CUB-200 data sets.}
% 		\vspace{-6pt}
		\label{tab:fidelity}
\end{table}

% \begin{table}[h!]
% \footnotesize
%     \centering
% \begin{tabular}{|p{1.45cm}| p{1.2cm} p{1.2cm}|p{1.1cm}|p{0.9cm}|}
% 		\hline \hline
% 			& \multicolumn{2}{|c|}{Baselines}&\multicolumn{2}{|c|}{OURS}\\
% 			Dataset & LIME \cite{} & VIBI \cite{} & w/o sup & w sup \\
% 			\hline \hline
%             CIFAR10 & 31.5 & 65.5 & \textbf{99.11} & NA \\
%             \hline
%             ImageNet  & - & - & \textbf{?} & NA \\
%             \hline
%             AwA2     & - & - & 97.03 & \textbf{97.52} \\
%             \hline
%             CUB-200 & - & - & 93.21 & \textbf{94.52} \\
%             \hline \hline
% 		\end{tabular}
% 		\caption{Comparison of fidelity metric of different methods on CIFAR10, ImageNet, AwA2 and CUB-200 datasets using ResNet18 architecture for concept encoder}
% % 		\vspace{-6pt}
% 		\label{tab:fidelity}
% \end{table}

\noindent \textbf{Explanation Error}: In data sets like CUB and AwA2, where ground truth concepts are available, we also measure how close are the concepts learned to the ground truth. We compute the $L_2$ distance between the concepts learned and the ground truth concepts to measure the alignment. From table~\ref{tab:alignment}, we can observe that the concepts generated by our method is most aligned to the ground truth concepts. While this should be the case for methods with concept supervision, our method without concept supervision also performs better than SENN, which illustrates our method's effectiveness in learning concept-based explanation even when annotations for concepts (or attributes) aren't available.

\begin{table}[h!]
\footnotesize
    \centering
    \begin{tabular}{|p{2.90cm}|p{1.50cm}|p{1.50cm}|}
		\hline \hline
			Dataset & AWA2 & CUB \\
			\hline \hline
            SENN & 0.99 & 1.34 \\
            \hline
            CBM & 0.91 & 1.17 \\
            \hline
            OURS (w/o sup) & 0.97 & 1.29 \\
            \hline
            OURS(w sup) & \textbf{0.89} & \textbf{1.14} \\
            \hline
            \hline
		\end{tabular}
% 		\label{tab:Flower main}
	    \vspace{-8pt}
		\caption{Comparison of \textit{explanation error} (we measure the mismatch using $L_2$ distance, hence lower the better) between concepts generated by different methods and the ground truth concepts (or attributes) on AwA2 and CUB-200 data sets.}
		 \vspace{-8pt}
		%\vspace{-10pt}
		\label{tab:alignment}
\end{table}

% average accuracy of predicting each ground truth concept using the learnt one that predicts it best. \cite{yeh2019completeness}

\noindent \textbf{Intervention on Concepts}: To study the concepts' usefulness, we scale their values in the $[0,1]$ range, select those above the threshold value $\omega$, set the concepts to 0, and then predict the label solely based on the intervened concept vector.  A change in the prediction means that the concepts zeroed are essential for explaining the model's decision. We repeat this procedure for all the instances in the test set and measure the predictive performance solely based on the generated concepts. A lower value indicates that the concepts generated are faithfully explaining the models' decisions. Ideally, the predictive ability of the concepts generated by methods like SENN and CBM should be higher. Since, in their cases, the interpreter (or the explainer) is directly used to generate the model prediction, the predictive performance based on concepts generated should be lower. But, you can observe from table \ref{tab:intervention} the predictive performance after the intervention is the lowest for the proposed framework. 

\begin{table}[h!]
\footnotesize
    \centering
\begin{tabular}{|p{1.45cm}| p{1.2cm} p{1.2cm}|p{1.1cm}|p{0.9cm}|}
		\hline \hline
			& \multicolumn{2}{|c|}{Baselines}&\multicolumn{2}{|c|}{OURS}\\
			Dataset & SENN & CBM & w/o sup & w sup \\
			\hline \hline
            CIFAR10 & 66.57 & NA & \textbf{43.19} & NA \\
            \hline
            ImageNet  & 43.91 & NA & \textbf{34.52} & NA \\
            \hline
            AwA2     & 61.39 & 40.29 & 37.61 & \textbf{35.92} \\
            \hline
            CUB-200 & 47.22 & 36.11 & 34.38 & \textbf{32.59} \\
            \hline \hline
		\end{tabular}
		\caption{The effect of interventions (accuracy in \% after intervention, lower the better) on concepts generated by different methods for CIFAR10, ImageNet, AwA2 and CUB-200 data sets.}
% 		\vspace{-6pt}
		\label{tab:intervention}
\end{table}

\begin{figure*}
    \centering
    \includegraphics[scale=0.6]{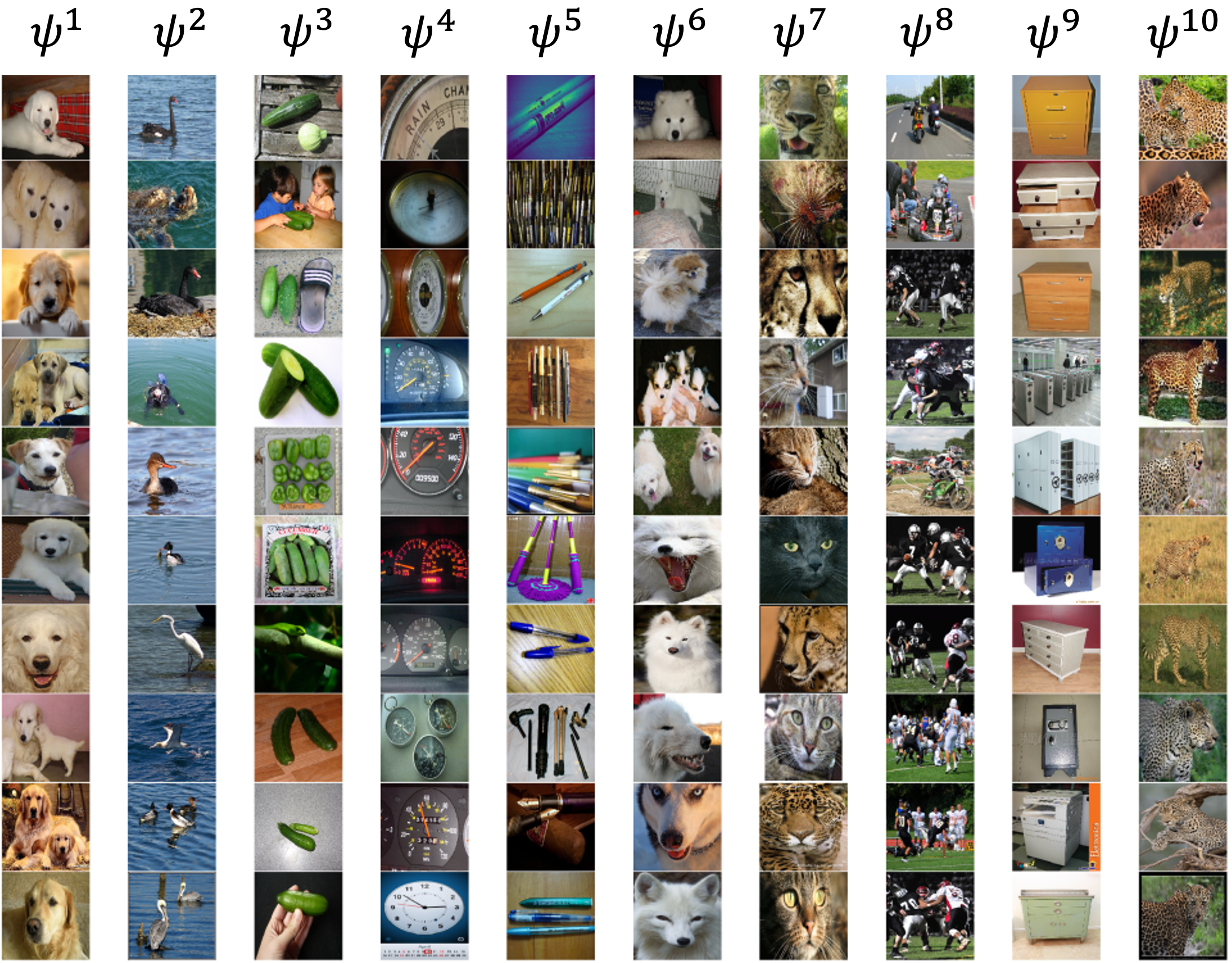}
    % \vspace{-3pt}
    \caption{A subset of 10 concept activations learnt by our framework on ImageNet. All these examples were correctly predicted by the model, and it can be seen that the each concept captures a certain set of homogeneous properties corresponding to a class. For ImageNet, we observe that the learned concepts are shared across the classes. For instance, $\psi^7$ is shared between tiger, cheetah, and different types of cat classes, and $\psi^6$ is shared among different forms of wolf and dog classes.}
    % \vspace{-8pt}
    \label{fig:concepts_imagenet}
\end{figure*}

\begin{figure*}
    \centering
    \includegraphics[scale=0.4]{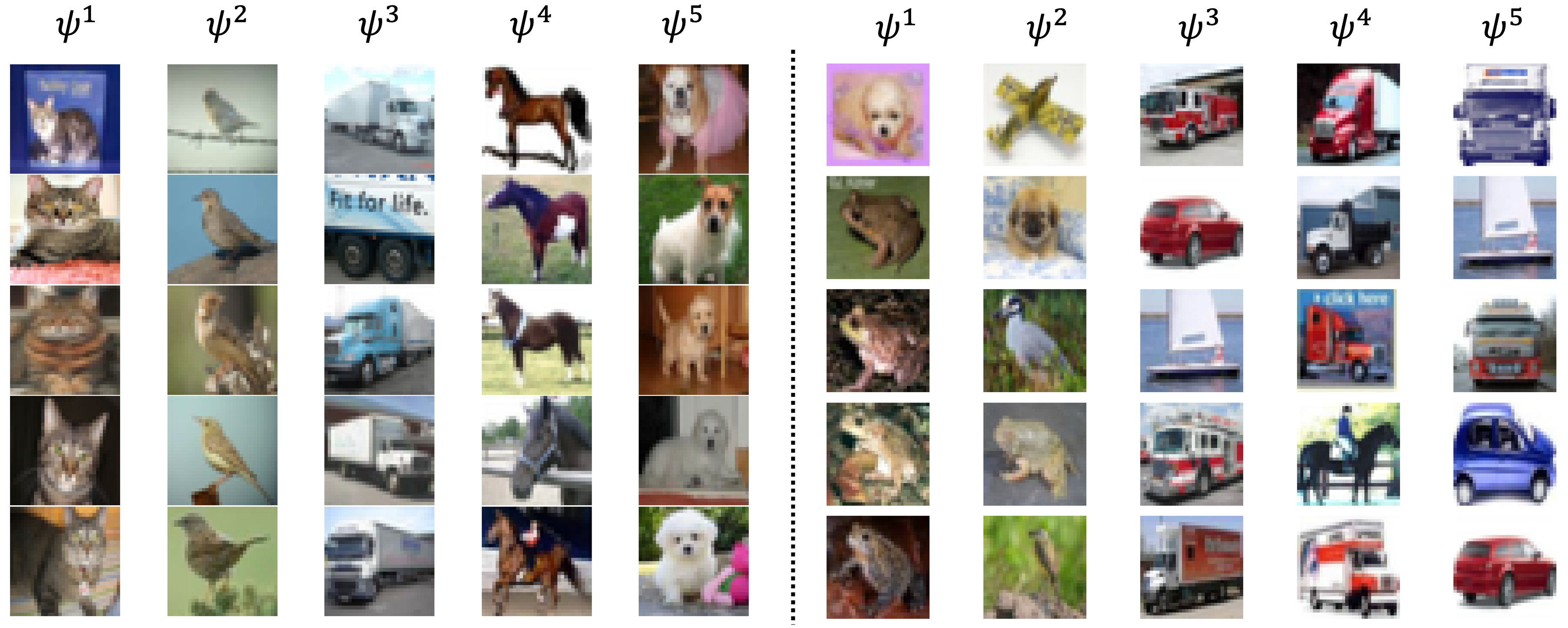}
    % \vspace{-3pt}
    \caption{Effect of decoder for CIFAR10 dataset. We can see the without the decoder \& the corresponding reconstruction loss the concept-based explanations (on the right) doesn't capture any homogeneous property and are hard to understand, unlike the case where decoder is present (on the left)}
    % \vspace{-8pt}
    \label{fig:concepts_cifar_decoder}
\end{figure*}

\vspace{-10pt}
\subsection{Qualitative Results}
\vspace{-5pt}
Qualitative results are significant for methods that explain models through concept-based representations. We generate explanations corresponding to every concept as the most representative images from the dataset. We present results from CIFAR10 and ImageNet datasets in the main paper and move the rest to the Appendix due to space constraints. The top concept activations generated for ImageNet are presented in fig.\ref{fig:concepts_imagenet}. We can observe that every concept captures homogeneous characteristics from the dataset that mostly corresponds to a class or similar class type. For example, $\psi^7$ represents concepts of faces for cheetah and some other similar types of cat species.

For data sets like CIFAR10, where there isn't much intersection of higher-level attributes across classes, we observe that each learned concept only corresponds to characteristics (or features) from a single class. In comparison, for data sets like ImageNet, where there is a lot of intersection between the higher-level attributes of different classes, we observe that the learned concepts are shared across the classes. For instance, $\psi^7$ is shared between tiger, cheetah, and different types of cats classes, and $\psi^6$ is shared among different types of wolf and cat classes (refer figs~\ref{fig:concepts_imagenet} \& \ref{fig:concepts_cifar_decoder}).

\subsection{Global Explanations}
\begin{figure}
  \centering
  \includegraphics[scale=0.5]{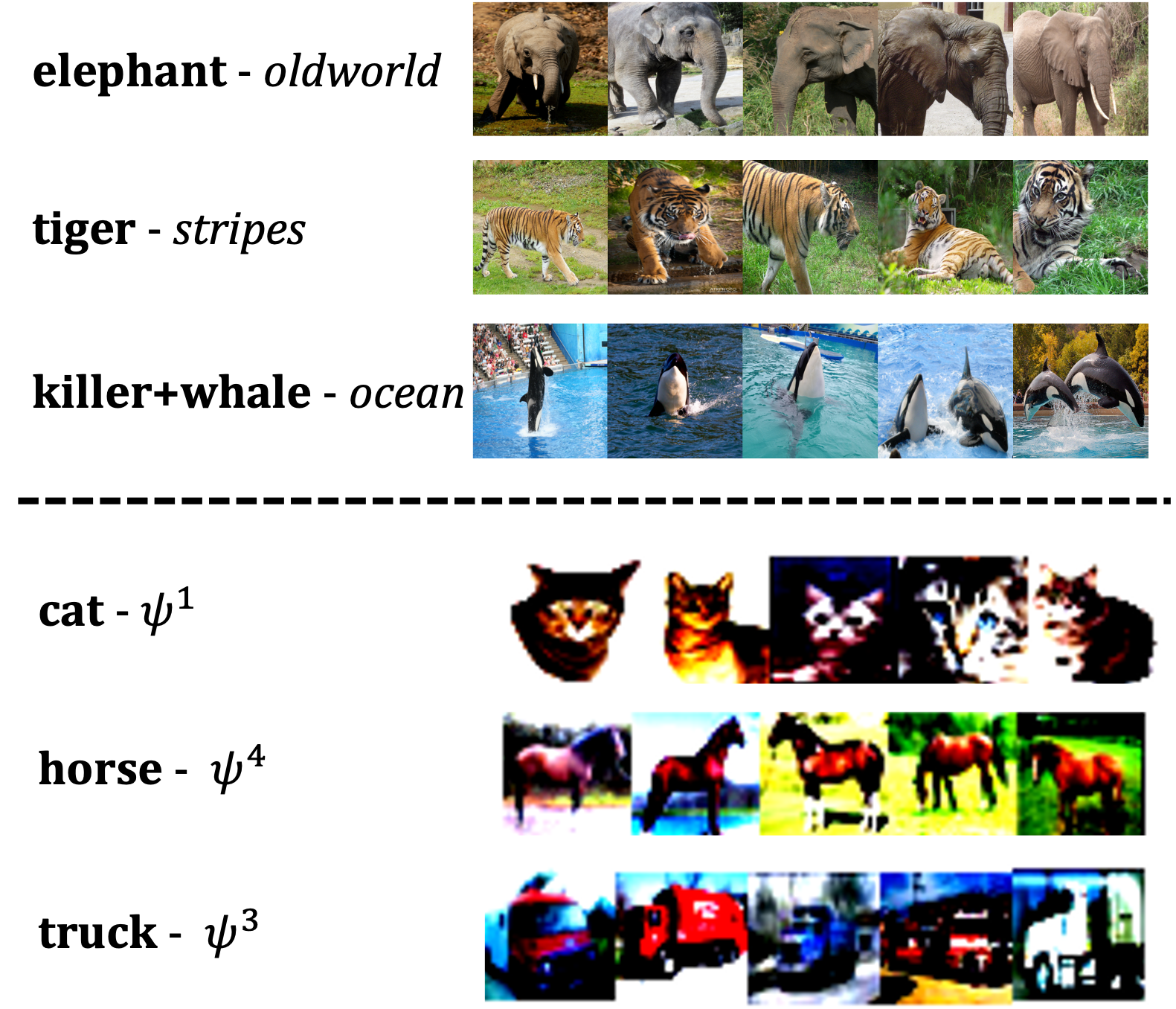}
  \caption{Example class-attribute pair analysis on AwA2 and CIFAR10 datasets with high global relevance (proportion of co-occurrence)}
  \label{fig:global}
\end{figure}

\vspace{-10pt}
An advantage of concept-based explanation methods compared to others is that they provide local as well global explanations. We identify class-concept (or attribute) pairs with a high proportion of co-occurrence to generate global explanations. We consider CIFAR10 and AwA2 for our experiments to explain the effectiveness of our method in generating such global explanations on datasets without and with ground truth concepts. 
% Images from a class is selected 
Simply analyzing these can reveal helpful information about the generated concepts. 
For instance, based on samples, we can see that (from fig.\ref{fig:global}) concept 'ocean' is a distinguishing attribute for class killer+whale of AwA2. Similarly, $\psi^1$ represents a distinctive concept for cat class of CIFAR10 (from $\psi^1$ to $\psi^5$ of fig.\ref{fig:concepts_cifar_decoder} on the left).

\vspace{-5pt}
\section{Ablation Studies}
\vspace{-5pt}
\label{sec:ablation}
% The following section presents different ablation studies to illustrate the importance of additional components incorporated by our framework.

\noindent \textbf{Importance of Self-supervision}:
As discussed in Sec.\ref{sec:method}, our framework enables us to incorporate self-supervision on the set of concepts easily, and this helps us to improve the quality of concepts by leveraging the underlying structure of the data. Our experiments use rotation prediction as the auxiliary self-supervision task for CIFAR10 and ImageNet datasets due to the unavailability of ground truth concepts. To compare the quantitative and qualitative performance with unsupervised concept training, we train our model with self-supervision for AwA2 and CUB datasets too. Table \ref{tab:predictive-performance-self-sup} reports the predictive performances of our method (with self-supervision on concepts and without any supervision) and SENN, as these methods don't require any ground truth concepts. Please note that the self-supervision is performed only on the concepts. Hence, this doesn't improve the predictive performance but rather enhances the faithfulness of the learned concept-based explanations. From table~\ref{tab:faithfulness-selfsup}, we can observe that self-supervision is improving the predictive performance through the concepts, and this, in turn, validates our hypothesis that leveraging the underlying structure of the data can the quality of concepts.

\begin{table}
\footnotesize
    \centering
\begin{tabular}{|p{1.45cm}| p{1.2cm} |p{1.1cm}|p{1.4cm}|}
		\hline \hline
			& \multicolumn{1}{|c|}{Baseline}&\multicolumn{2}{|c|}{OURS}\\
			Dataset & SENN  & w/o sup & w self-sup \\
			\hline \hline
            CIFAR10 & 84.5 & 91.68 & 91.28 \\
            \hline
            ImageNet  & 58.55 & 65.09 & 64.84 \\
            \hline
            AwA2     & 76.41 & 81.04 & 79.89 \\
            \hline
            CUB-200 & 58.81 & 63.05 & 61.93 \\
            \hline \hline
		\end{tabular}
		\caption{Comparison of predictive performance (accuracy in \%) of models on different methods on CIFAR10, ImageNet, AwA2 and CUB-200 data sets with and without self-supervision}
		\vspace{-6pt}
		\label{tab:predictive-performance-self-sup}
\end{table}

\begin{table}[h!]
\footnotesize
    \centering
\begin{tabular}{|p{1.45cm}| p{1.2cm} |p{1.1cm}|p{1.4cm}|}
		\hline \hline
			& \multicolumn{1}{|c|}{Baseline}&\multicolumn{2}{|c|}{OURS}\\
			Dataset & SENN & w/o sup & w self-sup \\
			\hline \hline
            CIFAR10 & 84.50 & 90.86 & 90.93 \\
            \hline
            ImageNet  & 58.55 & 58.73 & 60.28 \\
            \hline
            AwA2     & 76.41 & 79.29 & 79.77 \\
            \hline
            CUB-200 & 58.81 & 61.49 & 61.81 \\
            \hline \hline
		\end{tabular}
		\caption{Comparison of \textit{faithfulness (in \%)} (predictive capacity of the generated concepts) of concepts generated by different methods on CIFAR10, ImageNet, AwA2 and CUB-200 data sets with and without self-supervision.}
% 		\vspace{-6pt}
		\label{tab:faithfulness-selfsup}
\end{table}

\noindent \textbf{Importance of Reconstruction Error}:
A decoder improves the quality of the concepts by enforcing sufficiency, i.e., making them capable of faithfully reconstructing the image. In other words, this compels the set of concepts to capture all the image information and make the set of concepts complete. We measure the effect of the decoder by training our model for the CIFAR10 dataset without the decoder, keeping all the other model parts unchanged. We generate the explanations of the trained model and present them in fig. \ref{fig:concepts_cifar_decoder}. For comparing with the complete model (i.e., our model with decoder), we add explanations generated by our complete model in the same figure. The first and the last five columns are explanations generated by our complete model and the model without a decoder, respectively. These examples support our claim about the importance of decoder for learning better concepts. Please note that the model without a decoder performs slightly better than our complete model, but sacrificing a little bit of predictive performance can be justified to gain trust in the model.

% \subsection{Effect of Concept Removal During Training}

\vspace{-10pt}
\section{Conclusion}
\vspace{-10pt}
\label{sec:conclusion}
 In this work, we propose a new framework towards learning ante-hoc concept-based explanations that: (i) can be added easily to existing backbone classification architectures with minimal additional parameters; (ii) can provide explanations for model decisions in terms of concepts for an individual input image or groups of images; \& (iii) can work with different levels of supervision, including no concept-level supervision at all. Even though our framework incorporates additional components to existing deep learning backbones (or pipelines), we can discard most of them after the training. We only retain the sub-network (or module) to generate explanations in addition to the ones on standard deep learning pipelines (i.e., feature extractor and the classifier function) during the prediction time. Hence, compared to existing self-explaining models, the additional cost incurred by our framework is relatively insignificant. We performed a comprehensive suite of experiments to study the accuracy and explainability of our method on multiple benchmark datasets both quantitatively and qualitatively. Our approach consistently outperforms the baseline methods in all the datasets. In addition to this, we also performed ablation studies to illustrate the importance of additional components added by our method.

% \paragraph{Limitations \& Broader Impacts}: 

{\small
\bibliographystyle{ieee_fullname}
\bibliography{main}
}

\clearpage

\appendix
\section*{Supplementary: A Framework for Learning Ante-hoc Explainable Models via Concepts}

\section*{Limitations \& Broader Impact}
\label{sec:limitations}
The increasing need for explainability in the use of Deep Neural Network (DNN) models has in turn necessitated the design of ante-hoc explainable models that jointly learn to predict and explain. The limited efforts in this space such as SENN~\cite{alvarez2018towards} and CBM~\cite{koh2020concept} have their own set of limitations when used in practice. They either require concept-level supervision to train the model or need a significant number of additional parameters in the network, which prohibits their use in deeper models more commonly used in practice. Our framework aims to alleviate these issues. The proposed work addresses this need, and provides a framework for learning ante-hoc explainable models via concepts with significantly lesser additional parameters when compared to ~\cite{alvarez2018towards} (please see Sec \ref{sec:expts} of main paper). 

A limitation of our method is this increase in computational cost during training due to adding additional components such as the decoder network and the concept generator to an existing DL pipeline (although this is still better than baseline methods such as SENN). We keep additional components like the decoder network in the model despite this limitation since the decoder encourages the interpreter (or the concept generator) to generate meaningful explanations and faithfully capture semantics of an input image. While our model needs additional training time, its inference time, which matters in practice, isn't significantly higher than the backbone architecture itself. Our framework works by jointly learning to generate ante-hoc explanations via concepts and predict the label for the given image. Ante-hoc methods like ours can help understand a model's decision and gain intuition into its inner workings. In turn, this can help to improve the transparency and trustworthiness of Deep Learning models.  In addition to the above, our framework also allows a user to intervene on learned concepts to understand a model's decisions. In turn, this could be a valuable tool, especially when models are deployed for sensitive or critical tasks.

\section*{Appendix}
\label{sec:appendix}
\vspace{-5pt}
In this appendix, we provide details that we could not include in the main paper owing to space constraints, including:
\vspace{-5pt}
\begin{itemize}
\setlength\itemsep{-0.05em}
    \item Comparison of the faithfulness of explanations generated by our technique against different post-hoc explainability methods such as LIME and Grad-CAM
    \item Predictive performance of our framework with different backbone architectures
    \item Comparison of concept-based explanations generated by our method with and without concept supervision, in cases where annotations of attributes are available
    \item  Comparison of concept-based explanations generated by our framework and baselines methods such as SENN and CBM
    \item Comparison of concept-based explanations generated by our framework with different number of concepts.
\end{itemize}

\section{Comparison with Post-hoc Methods} 
In the recent past, there has been a lot of interest in developing techniques that try to explain a model's prediction after training. While these methods are helpful, the separation of explanation from prediction is not ideal. Ideally, we would like the techniques which generate interpretations to explain the model's prediction faithfully. But in the case of post-hoc methods, when an explanation goes wrong, it is not trivial to understand if the explanation method is incorrect or if the model itself relied on spurious correlations to make a prediction. To illustrate that our framework generates more faithful explanations than post-hoc methods, we compare the \textit{faithfulness} (predictive capacity of the generated concepts, i.e., from the output of $s_{\theta_{cce}}$) of well-known post-hoc explainability methods such as LIME~\cite{attr2016perturb_3_lime} and Grad-CAM~\cite{grad-cam} with that of our framework. We generate explanations with these methods for every image and pass the modified image through the model. From Table \ref{tab:fidelity post-hoc}, we see that our method outperforms other post-hoc explainability methods in terms of the faithfulness metric.

\begin{table}[h!]
\footnotesize
    \centering
\begin{tabular}{|p{1.45cm}| p{1.2cm} p{1.2cm}|p{1.1cm}|p{0.9cm}|}
		\hline \hline
			& \multicolumn{2}{|c|}{Post-hoc Baselines}&\multicolumn{2}{|c|}{OURS}\\
			Dataset & LIME & Grad-CAM & w/o sup & w sup \\
			\hline \hline
            CIFAR10 & 34.23 & 76.91 & \textbf{90.86} & NA \\
            \hline
            ImageNet  & 29.39 & 47.48 & \textbf{59.73} & NA \\
            \hline
            AwA2     & 46.21 & 75.00 & 79.29 & \textbf{83.30} \\
            \hline
            CUB-200 & 27.68 & 43.51 & 61.49 & \textbf{62.59} \\
            \hline \hline
		\end{tabular}
		\caption{Comparison of \textit{faithfulness} (predictive capacity of the generated concepts) our method against post-hoc explainability methods on CIFAR10, ImageNet, AwA2 and CUB-200 data sets. (w/o sup = without supervision; w sup = with supervision)}
% 		\vspace{-6pt}
		\label{tab:fidelity post-hoc}
\end{table}

\section{Plug and Play: Integrating with Other Backbone Architectures} 
One of the advantages of our method is that we can plug ante-hoc interpretability to different existing DNN backbone architectures. To illustrate this, we incorporate ante-hoc interpretability for 4 different popular backbones architectures i.e. ResNet34, ResNet50, EfficientNet-B0 and DenseNet-121 in Table~\ref{tab:different architecture}. We consider one dataset each with and without concept supervision for our experiments to show these results, i.e.  and CIFAR10 and AwA2 respectively. We find that a stronger backbone (such as DenseNet-121) helps improve the performance of our framework.

\begin{table}[h!]
\footnotesize
    \centering
    \begin{tabular}{|p{2.90cm}|p{1.50cm}|p{1.50cm}|}
		\hline \hline
			Architecture & CIFAR10 & AwA2 \\
			\hline \hline
            ResNet34 & 91.82 & 85.88 \\
            \hline
            ResNet50 & 92.04 & 86.11 \\
            \hline
            EfficientNet-B0 & 91.79 & 85.95 \\
            \hline
            DenseNet-121 & \textbf{92.85} & \textbf{86.73} \\
            \hline
            \hline
		\end{tabular}
% 		\label{tab:Flower main}
	    \vspace{-8pt}
		\caption{Comparison of \textit{accuracy} (in \%) on CIFAR10 and AwA2 datasets using different backbone architectures as concept (or base) encoder. All numbers with AwA2 are generated with concept supervision.}
		 \vspace{-8pt}
		%\vspace{-10pt}
		\label{tab:different architecture}
\end{table}

\section{Qualitative Evaluation} 
This section presents additional qualitative studies on the quality of concepts and an ablation study on change in number of concepts.

\subsection{Quality of Concepts}
\begin{figure*}
    \centering
    \includegraphics[scale=0.6]{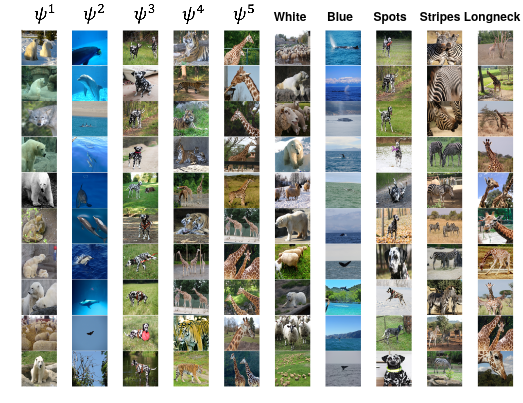}
    % \vspace{-3pt}
    \caption{A subset of 5 concepts learned by our framework on AwA2 each with (right) and without (left) concept supervision. Please note that some of the concepts learned without concept supervision capture similar concepts that are learned with concept supervision.}
    % \vspace{-8pt}
    \label{fig:concepts_awa_different_supervision}
\end{figure*}

\begin{figure*}
    \centering
    \includegraphics[scale=0.6]{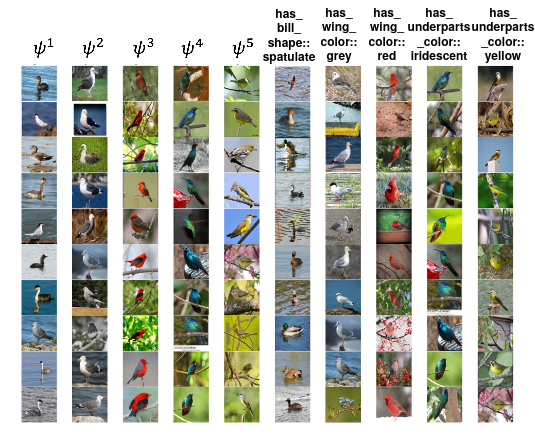}
    % \vspace{-3pt}
    \caption{A subset of 5 concepts learned by our framework on CUB-200 each with (right) and without (left) concept supervision. We observe that some of the concepts learned without concept supervision capture similar concepts that are learned with concept supervision.}
    % \vspace{-8pt}
    \label{fig:concepts_cub_different_supervision}
\end{figure*}

\noindent \textbf{Concepts: \textit{With and Without Supervision}}:
In this section, we observe that the concepts discovered by our framework correspond to some of the ground truth attributes even when trained without concept supervision. We considered AwA2 and CUB-200 for this study, as these datasets have ground-truth concepts for every image. We generated a set of images that maximally activate each concept for both types of models, i.e., with and without concept supervision, and present such results for some of the concepts in Figures \ref{fig:concepts_awa_different_supervision} and \ref{fig:concepts_cub_different_supervision} for AwA2 and CUB-200 respectively. In these figures, five concepts on the left are generated from the model trained without concept supervision. The rest of the concepts were generated by the model trained with concept supervision.  For example, the images for $\psi^5$ (from the model without concept supervision) are visually similar to the images for the \textsc{longneck} concept (from the model with concept supervision) from Figure \ref{fig:concepts_awa_different_supervision}. Also, the images for $\psi^3$ (from the model without concept supervision) are visually similar to the images for \textsc{has\_wing\_color::red} concept (from the model with concept supervision) from Figure \ref{fig:concepts_cub_different_supervision}.

\begin{figure*}
    \centering
    \includegraphics[scale=0.55]{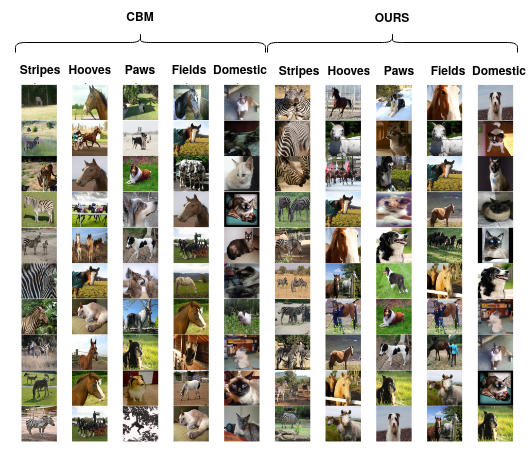}
    % \vspace{-3pt}
    \caption{A subset of 5 concepts learned by our framework on AwA2 each with CBM (left) and our method (right). Here we consider our method with concept supervision for fair comparison with CBM.}
    % \vspace{-8pt}
    \label{fig:concepts_awa_cbm_ours}
\end{figure*}

\begin{figure*}
    \centering
    \includegraphics[scale=0.55]{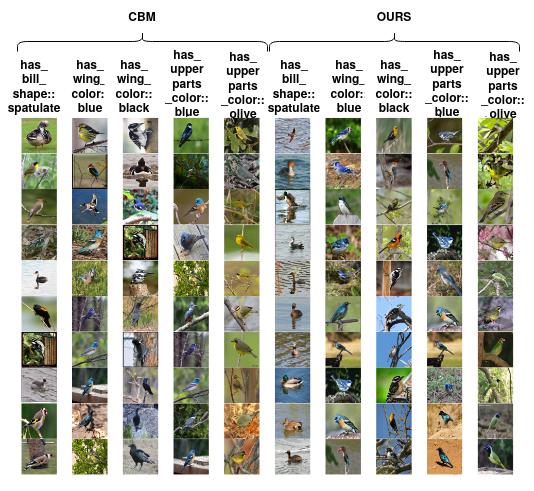}
    % \vspace{-3pt}
    \caption{A subset of 5 concepts learned by our framework on CUB-200 each with CBM (left) and our method (right). Here we consider our method with concept supervision for fair comparison with CBM.}
    % \vspace{-8pt}
    \label{fig:concepts_cub_cbm_ours}
\end{figure*}

\vspace{9pt}
\noindent \textbf{Concepts: \textit{Ours vs CBM}}:
Apart from the quantitative results presented in the main paper, which compare our method with the baseline method that considers concept supervision, i.e., Concept Bottlencek Methods (CBM), we herein show sample qualitative results. We generated images that maximally activate every concept learned by our model as well as CBM and show in Figures \ref{fig:concepts_awa_cbm_ours} and \ref{fig:concepts_cub_cbm_ours} for AwA2 and CUB-200 respectively. While all the concepts are meaningful visually, we can see a better selection of representative images for the concepts generated by our model. For example, images activated for the \textsc{fields} concept for our model represent the concept better than the images activated for CBM from fig.\ref{fig:concepts_awa_cbm_ours}. Similarly, images activated for \textsc{has\_bill\_shape::spatulate} concept for our model represent the concept better than the images activated for CBM from Figure \ref{fig:concepts_cub_cbm_ours}.

\begin{figure*}
    \centering
    \includegraphics[scale=0.65]{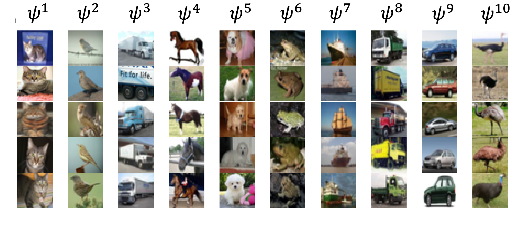}
    \vspace{-16pt}
    \caption{Concept activations (i.e. images that maximally activate each concept) learned by our framework on CIFAR10. It can be seen that each concept captures a certain set of homogeneous properties corresponding to a class. For instance, $\psi^1$ is mostly activated for images from cat class and the same for $\psi^6$ happens for images from frog class.}
    \vspace{-12pt}
    \label{fig:concepts_cifar_ours}
\end{figure*}

\begin{figure*}
    \centering
    \includegraphics[scale=0.65]{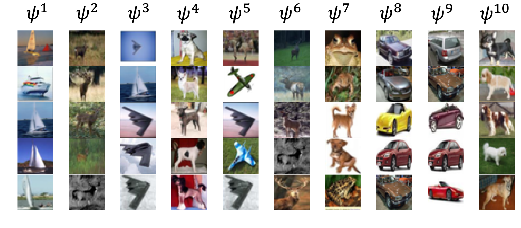}
    \vspace{-16pt}
    \caption{Concept activations (i.e. images that maximally activate each concept) learned by SENN on CIFAR10. It can be seen that some of the concepts are not able to capture a certain set of homogeneous properties corresponding to a class. For instance, $\psi^5$ is mostly activated for images from aeroplane class along with one image from horse class. Also, the maximally activated images for $\psi^5$ are from frog and dog classes.}
    \vspace{-12pt}
    \label{fig:concepts_cifar_senn}
\end{figure*}

\begin{figure*}
    \centering
    \includegraphics[scale=0.6]{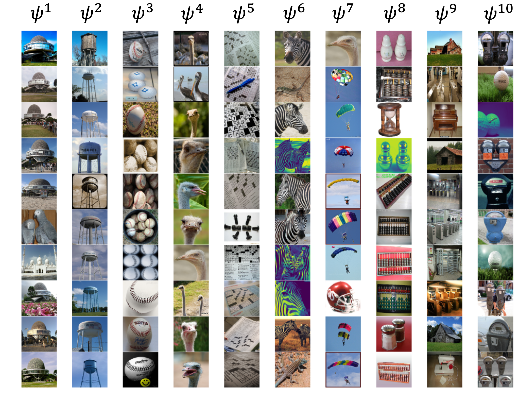}
    \vspace{-10pt}
    \caption{Concept activations (i.e. images that maximally activate each concept) learned by SENN on ImageNet. It can be seen that some concepts are not able to capture a certain set of homogeneous properties corresponding to one or many classes. For instance, $\psi^8$ is mostly activated for images which are visually not of similar structure. Also, most of the concepts capture a few of the properties, thus miss out on other important artifacts. For instance, $\psi^1$, $\psi^2$, $\psi^3$ and $\psi^{10}$ capture round shaped objects or objects with round shaped head.}
    \vspace{-18pt}
    \label{fig:concepts_imagenet_senn}
\end{figure*}

\vspace{9pt}
\noindent \textbf{Concepts: \textit{Ours vs SENN}}:
We present more qualitative results to compare the concepts generated by our method (without concept supervision) and SENN for CIFAR10 and ImageNet datasets. Figures \ref{fig:concepts_cifar_ours} and \ref{fig:concepts_cifar_senn} represent results with CIFAR10 for our method (without concept supervision) and SENN respectively. We observe that the concepts captured by SENN tend to repeat more and are less diverse than those generated by our method, thus leaving out some important aspects of the dataset. Similar issues are observed from the concept activations generated by SENN for ImageNet in Figure \ref{fig:concepts_imagenet_senn}. For example, many concepts capture round-shaped objects or objects with a round head and miss some important hidden concepts in the dataset (for concept-based explanations generated by our method on ImageNet, please see Figure \ref{fig:concepts_imagenet}). 

\subsection{Ablation Study: Number of Concepts}
\begin{figure*}
    \centering
    \includegraphics[scale=0.5]{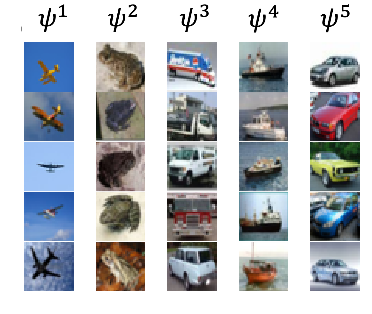}
    % \vspace{-3pt}
    \caption{Concept activations (i.e. images that maximally activate each concept) learned by our framework (with 5 concepts) on CIFAR10. It can be seen that each concept captures a certain set of homogeneous properties corresponding to a class. For instance, $\psi^1$ is mostly activated for images from aeroplane class and $\psi^2$ is mostly activated for images from frog class. It is also clear that all the important concepts are not captured by this model due to less number of concepts (i.e. 5 concepts).}
    % \vspace{-8pt}
    \label{fig:concepts_cifar_5_ours}
\end{figure*}

\begin{figure*}
    \centering
    \includegraphics[scale=0.65]{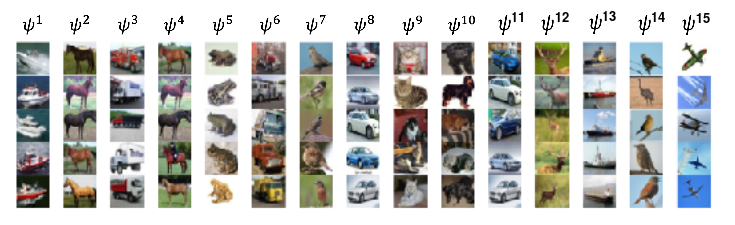}
    % \vspace{-3pt}
    \caption{Concept activations (i.e. images that maximally activate each concept) learned by our framework (with 15 concepts) on CIFAR10. It can be seen that each concept captures a certain set of homogeneous properties corresponding to a class. For instance, $\psi^7$ is mostly activated for images from bird class and $\psi^9$ is mostly activated for images from cat class. Also some of the concepts are repeated by this model due to more number of concepts (i.e. 15 concepts). For example, $\psi^2$, $\psi^4$ are both activated for images from horse class, $\psi^3$, $\psi^6$ are both activated for images from truck class and $\psi^8$, $\psi^{11}$ are both activated for images from automobile class.}
    % \vspace{-8pt}
    \label{fig:concepts_cifar_15_ours}
\end{figure*}

While the number of concepts is available apriori for the datasets with ground truth concepts, it's not known beforehand for other datasets like CIFAR10. Hence it is a choice (or hyperparameter) left to the user. To understand the impact of the number of concepts on the performance of our framework, we experimented with different number of concepts for CIFAR10 (i.e., 5 and 15) and achieve 91.51\% and 91.58\% accuracies, respectively. These numbers are very close to the model's accuracy with ten concepts (i.e., 91.68\%). We further analyzed the concepts generated by these two models (with 5 and 15 concepts) for more insights. We present the maximally activated images for every concept in Figures \ref{fig:concepts_cifar_5_ours} and \ref{fig:concepts_cifar_15_ours} for our models with 5 and 15 concepts respectively. It is evident that the model with 5 concepts (from Figure \ref{fig:concepts_cifar_5_ours}) is not able to capture all artifacts of the CIFAR10 dataset, whereas the model with 15 concepts (Figure \ref{fig:concepts_cifar_15_ours}) captures the most number of dataset artifacts (compared to models with 5 and 10 concepts), but it has repetitions of concepts. For example, the concept representing "deer" is not captured by any model except the model with 15 concepts. Exploring adaptive number of concepts and enforcing concept exclusivity while training could be interesting directions of future extensions of our work.

% \begin{figure*}
%     \centering
%     \includegraphics[scale=0.4]{Images/main_imagenet_concept_grid.png}
%     % \vspace{-3pt}
%     \caption{Concept activations learned by our framework on ImageNet.}
%     % \vspace{-8pt}
%     \label{fig:concepts_imagenet_ours}
% \end{figure*}


\end{document}